# Reliable Deep Diffusion Tensor Estimation:

# Rethinking the Power of Data-Driven Optimization Routine

Jialong Li[a,b,#], Zhicheng Zhang[a,c,#], Yunwei Chen[a,b], Qiqi Lu[a,b], Ye Wu[d], Xiaoming Liu[e], QianJin Feng[a,b], Yanqiu Feng[a,b,*], Xinyuan Zhang[a,b,*]

[a] School of Biomedical Engineering, Southern Medical University, Guangzhou, China
[b] Guangdong Provincial Key Laboratory of Medical Image Processing and Guangdong Province Engineering Laboratory for Medical Imaging and Diagnostic Technology, Southern Medical University, Guangzhou, China
[c] JancsiLab, JancsiTech, Hongkong, China
[d] School of Computer Science and Technology, Nanjing University of Science and Technology, Nanjing, China
[e] Department of Radiology, Union Hospital, Tongji Medical College, Huazhong University of Science and Technology, Wuhan 430022, China.

# Jialong Li and Zhicheng Zhang are co-first authors.
* Yanqiu Feng and Xinyuan Zhang are corresponding authors.
Correspondence to:
   Dr. Yanqiu Feng
   Email: foree@smu.edu.cn
   School of Biomedical Engineering, Southern Medical University, Guangzhou, China
and
   Dr. Xinyuan Zhang
   Email: zhangxyn@smu.edu.cn
   School of Biomedical Engineering, Southern Medical University, Guangzhou, China

# Abstract

Diffusion tensor imaging (DTI) holds significant importance in clinical diagnosis and neuroscience research. However, conventional model-based fitting methods often suffer from sensitivity to noise, leading to decreased accuracy in estimating DTI parameters. While traditional data-driven deep learning methods have shown potential in terms of accuracy and efficiency, their limited generalization to out-of-training-distribution data impedes their broader application due to the diverse scan protocols used across centers, scanners, and studies. This work aims to tackle these challenges and promote the use of DTI by introducing a data-driven optimization-based method termed DoDTI. DoDTI combines the weighted linear least squares fitting algorithm and regularization by denoising technique. The former fits DW images from diverse acquisition settings into diffusion tensor field, while the latter applies a deep learning-based denoiser to regularize the diffusion tensor field instead of the DW images, which is free from the limitation of fixed-channel assignment of the network. The optimization object is solved using the alternating direction method of multipliers and then unrolled to construct a deep neural network, leveraging a data-driven strategy to learn network parameters. Extensive validation experiments are conducted utilizing both internally simulated datasets and externally obtained in-vivo datasets. The results, encompassing both qualitative and quantitative analyses, showcase that the proposed method attains state-of-the-art performance in DTI parameter estimation. Notably, it demonstrates superior generalization, accuracy, and efficiency, rendering it highly reliable for widespread application in the field.

# 1. Introduction

Diffusion tensor imaging (DTI) serves to characterize the microstructure of biological tissue by measuring water molecule diffusion in a three-dimensional (3D) space through a second-order diffusion tensor (Basser et al., 1994b). Being among the most widely used magnetic resonance imaging (MRI) techniques, DTI has found extensive application in neuroscience and clinical research. Its uses include mapping developmental trajectories in the human brain, tracking fibers for brain tumor surgery planning, and early detection of various brain disorders (Mori and Zhang, 2006). However, the inherent low signal-to-noise ratio (SNR) on diffusion-weighted (DW) images undermines the reliability of estimated diffusion tensors and related parameters, such as fractional anisotropy (FA), mean diffusivity (MD), axial diffusivity (AD), and radial diffusivity (RD). This limitation, in turn, impacts scientific discoveries and clinical diagnoses. Hence, achieving reliable DTI parameter estimation is crucial to enhancing the practical application of DTI (Tax et al., 2021).

Theoretically, DTI necessitates acquiring a minimum of six DW images with non-collinear gradient directions and one non-DW image to estimate the diffusion tensor, which is a symmetric matrix with six independent unknown parameters for each voxel (Kingsley, 2006). Conventionally the least square algorithms are employed to fit the DW signals in a voxel-wise manner (Basser et al., 1994a; Collier et al., 2015; Guan Koay, 2010; Veraart et al., 2013). However, voxel-wise fitting proves highly sensitive to noise (Huang et al., 2015), demanding numerous DW images to counteract the severe noise effect. Guidelines suggest acquiring at least 30 non-collinear DW images (Jones, 2009), further extending scan times and risking motion artifacts. To enhance diffusion tensor quality from a limited number of DW images, various denoising algorithms have emerged. These methods fall into two categories: traditional denoising methods such as MPPCA (Veraart et al., 2016), GLHOSVD (Zhang et al., 2017b), NORDIC (Moeller et al., 2021), Patch2Self (Fadnavis et al., 2020), and deep learning-based methods such as DeepDTI (Tian et al., 2020), SDnDTI (Tian et al., 2022), and $DDM^2$ (Xiang et al., 2023). Despite the efficacy of "denoising-before-fitting" strategies in reducing noise effect, this decoupling of image denoising from parameter estimation leads to error accumulation in subsequent estimation stages.

Leveraging advancements in deep learning (DL), several deep learning-based methods have directly predicted diffusion parameters from noisy DW images. q-DL is one of the first deep learning-based methods for the estimation of diffusion parameters, in which a three-layer neural network was used to estimate diffusion kurtosis imaging (DKI) and neurite orientation dispersion and density imaging (NODDI) parameters using subsampled q-space imaging (Golkov et al., 2016). Subsequently, multilayer perceptron (Aliotta et al., 2021), convolutional neural network (CNN) (Karimi et al., 2021; Li et al., 2021), and transformer (Karimi and Gholipour, 2022) were employed to estimate DTI parameters from limited DW images. These methods of deep learning have demonstrated superior advantages such as accuracy and speed over conventional fitting methods. However, these methods, being purely data-driven, lack interpretability crucial for medical applications (Monga et al., 2021). Moreover, they are tailored for specific diffusion acquisition settings (e.g., b value, gradient directions), limiting their generalization to out-of-training-distribution datasets (Sabidussi et al., 2023).

Recently, model-based deep learning methods have emerged, combining optimized techniques with deep neural networks for various imaging inverse problems, particularly in medical image reconstruction (Aggarwal et al., 2019; Huang et al., 2023; Schlemper et al., 2018; Yang et al., 2020; Zhang et al., 2021a; Zhang et al., 2021b). These methods naturally incorporate prior structures and domain knowledge, unlike

"black-box" networks, resulting in better generalization and interpretability (Monga et al., 2021). Notably, only one study, dtiRIM (Sabidussi et al., 2023), utilized the model-based deep learning method for DTI. Here, recurrent inference machines learn a regularized solution while enforcing data consistency using the DTI model. dtiRIM exhibits good generalization to diverse acquisition protocols but relatively lower accuracy, especially with limited DW images.

To ensure accurate estimation of DTI parameters across various acquisition settings, we rethink the power of data-driven optimization and propose the DoDTI architecture, merging a well-structured model-based deep learning design. Specifically, the data fidelity term is constructed using the weighted linear least squares fitting (WLLS), while a deep learning-based denoiser forms the regularization term (Romano et al., 2017). This optimization function is then iteratively solved using the alternating direction method of multipliers (ADMM) algorithm. We have unrolled the ADMM's iterative steps into cascaded blocks, constructing an end-to-end deep network from DW images to diffusion tensor maps. In assessing DoDTI's generalization, we generated training datasets from only 16 Human Connectome project (HCP) subjects, each containing one non-DW image and six DW images sharing identical b values and gradient directions. Evaluation datasets encompass various clinical scenarios, including simulation datasets with different b values, gradient directions, SNRs, and spatially invariant/varying noise distribution, and in-vivo datasets from healthy subjects and patients across diverse acquisition settings at multiple centers. Both qualitative and quantitative analyses confirm that DoDTI excels in DTI parameter estimation (FA, MD, AD, and RD), demonstrating unparalleled accuracy, generalization, and speed, making it highly reliable for widespread applications.

## 2. Theory and Methods
### 2.1 Diffusion tensor model

In DTI, the noise-free DW signal $s$ for each voxel can be modeled by

$$s(b, \boldsymbol{g}_i) = s_0 e^{-b \boldsymbol{g}_i^{\mathrm{T}} D \boldsymbol{g}_i}, \tag{1}$$

where $b$ is the diffusion sensitivity factor, $\boldsymbol{g}_i = [g_{i1}, g_{i2}, g_{i3}]^{\mathrm{T}}$ is the column vector of the $i$-th gradient direction ($i = 1, \ldots, N$), $N$ is the total number of directions, $s_0$ is the non-DW signal (b = 0 s/mm$^2$), and $D = \{D_{ij}: i, j = 1, \ldots, 3\}$ is the second-order fully symmetric diffusion tensor.

### 2.2 WLLS Estimation

Because of the Rician distribution commonly followed by diffusion MRI (dMRI) signals, the typical nonlinear least squares (NLS) algorithm results in biased DTI estimation. The WLLS algorithm mitigates bias and enhances DTI accuracy by utilizing log-transformed dMRI signals and a weight term. WLLS has gained widespread use in DTI due to its simple implementation and minimal computational expense (Veraart et al., 2013).

For each voxel, the log-transformation of measured signals $\tilde{\boldsymbol{s}} = [\widetilde{s_0}, \tilde{s}(b, \boldsymbol{g}_1), \cdots, \tilde{s}(b, \boldsymbol{g}_N)]^{\mathrm{T}}$ can be expressed as

$$\boldsymbol{y} = A\boldsymbol{x} + \boldsymbol{\varepsilon}, \tag{2}$$

where $\boldsymbol{y} = \ln \tilde{\boldsymbol{s}}$, $\boldsymbol{x} = [\ln s_0, D_{11}, D_{22}, D_{33}, D_{12}, D_{23}, D_{13}]^{\mathrm{T}}$ contains seven unknown parameters to be estimated, $\boldsymbol{\varepsilon}$ denotes the column vector of independent error terms, $A$ is a mapping matrix designed by acquired b value and gradient directions as $A = [\boldsymbol{a}_0, \boldsymbol{a}_1, \cdots, \boldsymbol{a}_N]^{\mathrm{T}}$ where $\boldsymbol{a}_0 = [1, 0, 0, 0, 0, 0, 0]$, $\boldsymbol{a}_i = [1, -bg_{i1}^2, -bg_{i2}^2, -bg_{i3}^2, -2bg_{i1}g_{i2}, -2bg_{i2}g_{i3}, -2bg_{i1}g_{i3}]$.

Studies have shown that when the SNR of DW signals exceeds two, the expectation of $\boldsymbol{\varepsilon}$ becomes zero (Salvador et al., 2005). Under this condition, fitting log-transformed signals $\boldsymbol{y}$ yields an unbiased estimator of $x$. However, the variance of $\varepsilon$ is non-uniform and relies on noise-free signals $s$ as

$$\mathrm{var}(\varepsilon) = \left[\frac{\sigma^2}{s_0^2}, \frac{\sigma^2}{s^2(b, \boldsymbol{g}_1)}, \dots, \frac{\sigma^2}{s^2(b, \boldsymbol{g}_N)}\right], \tag{3}$$

where $\sigma$ is the noise level. The heterogeneity of error variance will reduce the precision of DTI parameter estimation. The WLLS is designed by using weights to correct the heterogeneity of error variance. Consequently, the WLLS was used to derive a closed-solution expressed as

$$\hat{x} = (A^\mathrm{T} W^\mathrm{T} W A)^{-1} A^\mathrm{T} W^\mathrm{T} W y, \tag{4}$$

where $W = \mathrm{diag}(s)$. Actually, the noise-free dMRI signals are unknown, and the weight matrix $W$ needs to be estimated.

### 2.3 Proposed method: DoDTI

Typically, diffusion tensors are independently estimated for each voxel, which becomes highly challenging due to low SNR and a limited number of DW images. The traditional model-based method, involving simultaneous estimation of all diffusion tensors across image space (referred to as diffusion tensor maps), incorporates prior information and has demonstrated improved accuracy in diffusion tensor estimation (Guo et al., 2022; Liu et al., 2013; Wang et al., 2004; Zhu et al., 2017). Recognizing that while deep learning-based methods excel in accuracy, they lack generalization compared to traditional model-based fitting techniques, we propose a **D**ata-driven **o**ptimization-based method for **DTI** (DoDTI). This method combines traditional WLLS with a deep learning-based denoiser as follows:

$$\hat{X} = \underset{X}{\mathrm{argmin}} \sum_{j=1}^{J} \frac{1}{2} \left\| W_j (A X_j - Y_j) \right\|_2^2 + \frac{\lambda}{2} \langle X, X - \mathcal{D}_\theta(X) \rangle, \tag{5}$$

where $j$ indexes the $j$-th voxel within the image domain, assuming the image size of $L_1 \times L_2 \times L_3$, with the total voxel count $J$. $W_j \in \mathbb{R}^{M \times M}$ is the weighting factor for $j$-th voxel, $M$ is the number of dMRI images, including $N$ DW images and $(M - N)$ non-DW images, mapping matrix $A \in \mathbb{R}^{M \times 7}$, $X = \{\ln S_0, D_{11}, D_{22}, D_{33}, D_{12}, D_{23}, D_{13}\} \in \mathbb{R}^{L_1 \times L_2 \times L_3 \times 7}$ is the unknown parameter maps, containing a log-transformed non-DW image and six unique tensor elements across the image domain, $X_j \in \mathbb{R}^{7 \times 1}$ represents the values of $X$ at $j$-th voxel, $Y \in \mathbb{R}^{L_1 \times L_2 \times L_3 \times M}$ is the log-transformed dMRI images, $\lambda$ is the regularization parameter, $\mathcal{D}_\theta(\cdot)$ is a 3D CNN-based denoiser with the trainable parameter $\boldsymbol{\theta}$, and $\langle \cdot \rangle$ denotes the inner product.

The first term, WLLS, ensures a zero expectation of fitting error $W_j(AX_j - Y_j)$ with a constant variance across image space and q-space. Moreover, it can fit DW images acquired from diverse acquisition protocols into parameter maps with a fixed number of channels, simplifying subsequent denoising procedures. The second term involves Regularization by Denoising (RED), which effectively regulates various inverse problems by integrating a denoising engine (Romano et al., 2017). The induced penalty is proportional to the inner product between the image and its denoising residual. The gradient of the RED term can be formulated as $\nabla_X \left( \frac{\lambda}{2} \langle X, X - \mathcal{D}_\theta(X) \rangle \right) = \lambda (X - \mathcal{D}_\theta(X))$, which avoids differentiating the denoiser function.

Using ADMM, Eq. (5) can be reformulated as a constrained optimization problem by introducing an auxiliary variable $Z$:

$$(\hat{X}, \hat{Z}) = \underset{X, Z}{\mathrm{argmin}} \sum_{j=1}^{J} \frac{1}{2} \left\| W_j (A X_j - Y_j) \right\|_2^2 + \frac{\lambda}{2} \langle Z, Z - \mathcal{D}_\theta(Z) \rangle \ \ s.t. \ X = Z. \tag{6}$$

Its augmented Lagrangian function is

$$\mathcal{L}_\rho(X, Z, \alpha) = \sum_{j=1}^{J} \frac{1}{2} \|W_j(AX_j - Y_j)\|_2^2 + \frac{\lambda}{2} \langle Z, Z - \mathcal{D}_\theta(Z) \rangle + \langle \alpha, X - Z \rangle + \frac{\rho}{2} \|X - Z\|_2^2, \quad (7)$$

where $\alpha \in \mathbb{R}^{L_1 \times L_2 \times L_3 \times 7}$ is the Lagrangian multiplier, $\rho$ is the penalty parameter. The scaled Lagrangian multiplier $\beta = \alpha/\rho$ ($\rho > 0$) is introduced for simplicity. The ADMM algorithm alternately optimizes $\{X, Z, \beta\}$ in each iteration by solving the following three subproblems:

$$\begin{cases} X^n = \underset{X}{\mathrm{argmin}} \sum_{j=1}^{J} \frac{1}{2} \|W_j(AX_j - Y_j)\|_2^2 + \frac{\rho}{2} \|X + \beta^{n-1} - Z^{n-1}\|_2^2 \\ Z^n = \underset{Z}{\mathrm{argmin}} \frac{\lambda}{2} \langle Z, Z - \mathcal{D}_\theta(Z) \rangle + \frac{\rho}{2} \|X^n + \beta^{n-1} - Z\|_2^2 \\ \beta^n = \beta^{n-1} + X^n - Z^n. \end{cases} \quad (8)$$

We unrolled the iterations of ADMM into a deep neural network with $N_s$ stages, as depicted in Fig. 1a. Each network stage corresponds to one ADMM iteration and comprises the fitting block (**X**), the auxiliary variable block (**Z**), and the multiplier block (**β**), addressing the three subproblems in Eq. (8), respectively. Subsequent sections will delve into a detailed description of each block.

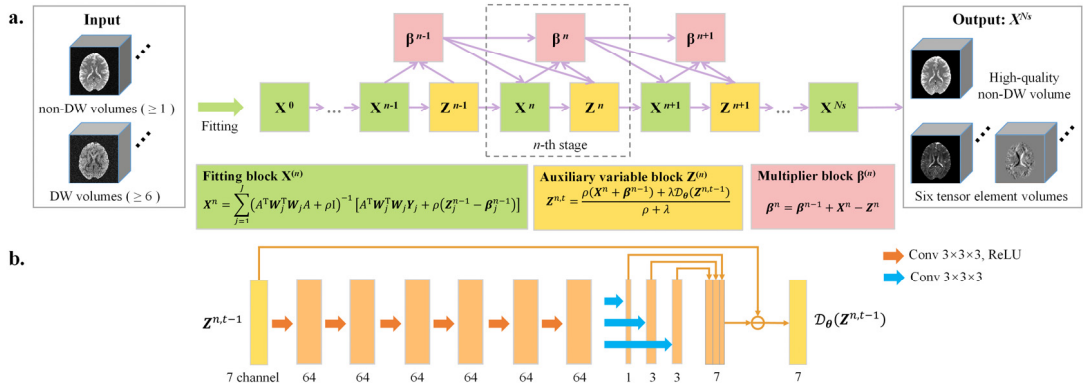

**Fig. 1.** Architecture of the proposed DoDTI. (a) The input is at least six DW volumes plus one non-DW volume and corresponding strength and directions of diffusion gradients. The output is the predicted non-DW volume and six tensor element maps. The n-th iteration of ADMM corresponds to the n-th stage of the deep neural network enclosed with the dashed box, in which the fitting block (**X**), the auxiliary variable block (**Z**), and the multiplier block (**β**) are sequentially updated. (b) Structure of the denoiser $\mathcal{D}_\theta(\cdot)$, which was embedded into the auxiliary variable block.

### 2.3.1 Fitting block (X)

The first subproblem in Eq. (8) can be solved using a close-form solution. Given $Z^{n-1}$ and $\beta^{n-1}$ from the previous stage, the output of the fitting block in *n*-th stage is defined as

$$X^n = \sum_{j=1}^{J} (A^\mathrm{T} W_j^\mathrm{T} W_j A + \rho I)^{-1} [A^\mathrm{T} W_j^\mathrm{T} W_j Y_j + \rho(Z_j^{n-1} - \beta_j^{n-1})], \quad (9)$$

where $I \in \mathbb{R}^{7 \times 7}$ is the identity matrix, and the weighting factor $W$ needs to be estimated, as noise-free dMRI images are unknown. The performance of DTI parameter estimation is highly dependent on the accuracy of $W$. To obtain accurate $W$, the estimated parameters $\widehat{X}$ from the previous stage were used to re-generate the dMRI images with the DTI model ($W_j = exp(A\widehat{X}_j)$), being a new weighting factor for the next stage.

### 2.3.2 Auxiliary variable block (Z)

The second subproblem shown in Eq. (8) is optimized using the fixed-point method (Chan et al., 2017). Given $X^n$ from the current stage and $\beta^{n-1}$ from the previous stage, the output of the $t$-th inner iteration of block $Z$ in the $n$-th stage is defined as

$$Z^{n,t} = \frac{\rho(X^n + \beta^{n-1}) + \lambda \mathcal{D}_\theta(Z^{n,t-1})}{\rho + \lambda}. \quad (10)$$

We initialize $Z^{n,0} = Z^{n-1}$, which is the output of block $Z$ in the $(n-1)$-th stage. The output in the $n$-th stage is $Z^n = Z^{n,N_t}$, where $N_t$ is the total number of inner iterations of block $Z$.

A 3D DnCNN $\mathcal{D}_\theta$ is integrated to aid the update of $Z^{n,t}$ by reducing noise and artifacts in $Z^{n,t-1}$. As shown in Fig. 1b, it employs a 7-layer CNN, where each layer incorporates $3 \times 3 \times 3$ kernels and ReLu activation function (except the final layer). For estimating parameter maps with distinct grayscale ranges $(\ln S_0); (D_{11}, D_{22}, D_{33}); (D_{12}, D_{23}, D_{13})$, three convolutional pathways are employed in the final layer. The DnCNN employs residual learning, known to be effective for image denoising and super-resolution (He et al., 2016). Note that $\rho$, $\lambda$, and $\theta$ are trainable parameters shared across different stages.

### 2.3.3 Multiplier block (β)

Given $X^n$ and $Z^n$ from the current stage, the output of the multiplier block in the $n$-th stage is defined as

$$\beta^n = \beta^{n-1} + X^n - Z^n. \quad (11)$$

### 2.3.4 Training loss

DoDTI was trained through supervised deep learning, with loss given by the expression

$$loss = \sum_{n=1}^{N_s} \frac{n}{N_s} (|X^n - X_{gt}| + |\mathcal{D}_\theta(Z^{n-1}) - X_{gt}| + |Z^n - X_{gt}|), \quad (12)$$

where $N_s$ is the total number of stages, $X_{gt} \in \mathbb{R}^{L_1 \times L_2 \times L_3 \times 7}$ is the ground truth parameter maps, and $X^n$, $\mathcal{D}_\theta(Z^{n-1})$ and $Z^n$, are the estimated maps at the $n$-th stage. The overall loss is composed of the mean absolute error of $X^n$, $\mathcal{D}_\theta(Z^{n-1})$ and $Z^n$, The weight ($\frac{n}{N_s}$) proportional to the stage index can guarantee a monotonic and fast convergence by guiding $X^n$, $\mathcal{D}_\theta(Z^{n-1})$ and $Z^n$ updated toward the ground truth at each stage.

## 3. Experiments
### 3.1 Model Training and Setting

We set $N_s = 8$, $N_t = 1$ to strike a balance between performance and computational cost. We set the initial values of $\rho$ and $\lambda$ as 0.001 and 0.1, respectively. The initial value of $X^0$ was determined through linear least squares fitting, assuming $Z^0 = X^0$, $\beta^0 = 0$. We employed a batch size of 4, trained for 250 epochs, with the learning rate initially set at 0.0001 and subsequently halved every 100 epochs. All experiments were conducted on a server equipped with an Intel Xeon E5-2667 CPU and an RTX2080 Ti GPU.

### 3.2 Dataset Details
#### 3.2.1 Training Data Generation

The dMRI data of 31 healthy subjects (16 subjects for training, 4 subjects for validation, and 11 subjects for testing) were obtained from the HCP MGH Adult Database (Fan et al., 2016). The HCP

dMRI dataset comprises 40 b = 0 s/mm² volumes, 64 b = 1000, 64 b = 3000, 128 b = 5000, and 256 b = 10000 s/mm² volumes for each subject. Note that our DTI study employed only the first five b = 0 s/mm² and 64 b = 1000 s/mm² dMRI data. The training data generation involved several steps: 1) denoising the dMRI data using MPPCA (Veraart et al., 2016); 2) removal of Gibbs artifacts and correction of intensity inhomogeneity; 3) estimation of DTI parameters (non-DW images ($S_0$) and diffusion tensor maps ($D_{ij}$) using WLLS, considered as the training label; 4) synthesis of six b = 1000 s/mm² DW images $\{S(1000, g_i), i = 1, \cdots, 6\}$ using estimated non-DW images ($S_0$) and diffusion tensor maps ($D_{ij}$) with the DTI model along the DSM diffusion directions scheme (Skare et al., 2000); 5) normalization of non-DW images and DW images using the 99-th percentile signal intensity of the non-DW image; 6) addition of Rician noise into the synthesized dMRI data $\{S_0; S(1000, g_i), i = 1, \cdots, 6\}$ to obtain the noisy dMRI data, serving as the input for training data. Sixteen evenly spaced values spanning from 0.005 to 0.045 were set as noise levels (σ) and each training data corresponds to one of the noise levels. Similarly, the noise levels for validation data were set to 0.01, 0.02, 0.03, and 0.04. Blocks of 32 × 32 × 32 voxels were employed for training.

### 3.2.2 Simulated testing data

Five experiments were designed to assess the generalization ability of DoDTI across variations in acquisition settings. The default acquisition setting is the same as that of the training data (b = 1000 s/mm², noise level $\sigma = 0.03$, one non-DW volume, and six DW volumes with DSM diffusion directions scheme). In each experiment, only one acquisition parameter was modified from its default value. As outlined in the section of training data generation, 11 simulated testing data instances were regenerated for each altered parameter setting.

- **Effect on different b values**: The b values include 800, 1000, and 1200 s/mm²;
- **Effect on varying six-direction settings**: The six-direction settings include the DSM scheme, as well as the DSM scheme rotated around the z-axis by 30, 60, and 90 degrees, and the Jones scheme (Jones et al., 1999);
- **Effect on different numbers of DW volumes**: The number of DW volumes includes 7, 15, 25, and 36;
- **Effect on different noise levels**: The noise level σ includes 0.01, 0.02, 0.03, and 0.04;
- **Effect on spatially varying noise**: The noise level increases linearly from the outside to the inside of the volume, ranging from 0.01 to 0.04.

### 3.2.3 In-vivo Data

The first in-vivo dataset was obtained from a published study (Moeller et al., 2021). The data were collected from a healthy volunteer on a 3T Siemens Magnetom Prisma scanner using a 32-channel head coil at the Center for Magnetic Resonance Research (CMRR), University of Minnesota. Data acquisition and image reconstruction were performed, with the CMRR distributed C2P multiband diffusion sequence. The detailed parameters were as follows: TR/TE = 3230/89 ms, matrix size = 140 × 140 × 92 with 1.5 mm isotropic resolution, the multi-band acceleration factor of 4, and phase partial Fourier = 6/8. A pair of dMRI data with the opposite encoding directions (AP and PA) was acquired, each consisting of seven b = 0 s/mm², 46 b = 1500 s/mm², and 46 b = 3000 s/mm² images, with six repeated scans. For diffusion tensor estimation, only one b = 0 s/mm² and 46 b = 1500 s/mm² were utilized in this study.

The second in-vivo dataset was acquired from an individual with white matter hyperintensities, while the third dataset originated from a patient with post-surgical removal of a glioma. Both datasets were acquired using a single-shot SE-EPI sequence on a 3T Philips Ingenia 3.0 CX scanner equipped with a 32-channel head coil at Wuhan Union Hospital, China. The dMRI data were gathered at two isotropic imaging resolutions of 1.5 mm and 2.0 mm, each containing one b = 0 s/mm² and 64 b = 1000 s/mm² images. Imaging parameters for the 1.5 mm isotropic data were: TR/TE = 2826/76 ms, matrix size = 160 × 160 × 96, while for the 2.0 mm isotropic data: TR/TE = 3000/79 ms, matrix size = 112 × 112 × 72.

The fourth in-vivo dataset was obtained from a glioma patient on a 3T Siemens Magnetom Prisma scanner equipped with a 64-channel head/neck coil at Tiantan Hospital, China (Guo et al., 2022). The CMRR multiband sequence was employed. Diffusion encoding was applied along 32 directions on b = 1000 s/mm² and 64 directions on b = 2000 s/mm². In addition, four b = 0 s/mm² images were acquired. The detailed imaging parameters were: TR/TE = 3500/98 ms, matrix size = 104 × 104 × 72 with 2 mm isotropic resolution, phase partial Fourier = 7/8, the multi-band acceleration factor of 3, and no in-plane parallel imaging acceleration. Only four b = 0 s/mm² and 32 b = 1000 s/mm² images were utilized for this study.

All in-vivo datasets underwent correction for eddy current and motion-induced distortion (Andersson et al., 2003; Andersson and Sotiropoulos, 2016) using FSL software (Jenkinson et al., 2012; Smith et al., 2004) (https://fsl.fmrib.ox.ac.uk/). All of these datasets were acquired after obtaining institutional review board approval and written informed consent.

### 3.3 Compared Methods

In our study, we conducted a comparative analysis involving our method and four other advanced techniques along with one widely adopted approach: (1) WLLS (Chung et al., 2006) from MRtrix (Tournier et al., 2019), a prevalent fitting method used for DTI parameter estimation; (2) MPPCA (Veraart et al., 2016) denoising, a noise reduction technique involving the projection of data onto its principal components, followed by WLLS, known as MPPCA-WLLS; (3) GL-HOSVD (Zhang et al., 2017b) denoising, which diminishes noise by decomposing high-order tensors, followed by WLLS, termed as GLHOSVD-WLLS; (4) DeepDTI (Tian et al., 2020) denoising, employing a DnCNN denoiser (Zhang et al., 2017a), followed by WLLS, referred to as DeepDTI-WLLS; (5) dtiRIM (Sabidussi et al., 2023), a DTI deep learning method known for its generalizability. Notably, in our comparison, DeepDTI was retrained using the same training data as in our work, and dtiRIM was directly tested using the model provided by its author.

### 3.4 Evaluation Metrics

To quantitatively assess the performance of these methods, we computed two key metrics: the normalized root mean square error (NRMSE) and the structural similarity index measure (SSIM) between the estimated and reference DTI parameters, encompassing FA, MD, AD, and RD for simulated data. Regarding the first three in-vivo data, high-quality parameter maps were obtained using WLLS with multiple repeated acquisitions and/or a high number of directions (≥64). These maps served as references for calculating NRMSE and SSIM, analogous to the simulation experiments. To expedite DTI acquisition time, we selected only six DW volumes from the complete dataset to estimate parameter maps. The chosen gradient directions aligned closely with those from the training data to ensure DeepDTI's performance. For the fourth in-vivo data, considering that 32 DW volumes are not large enough to produce the high-quality parameter maps as the reference, we use the coefficient of determination ($R^2$) to evaluate the consistency of maps obtained from 6 DW volumes and 32 DW volumes. Additionally, we employed fiber tracking using AFQ (Yeatman et al., 2012), facilitating deterministic tractography to reconstruct white-matter fascicle.

## 4. Results
### 4.1 Simulated Data Experiments
#### 4.1.1 Effect on different b values

Fig. 2 displays the NRMSE and SSIM of FA and MD estimated by various methods using simulated datasets with different b values. Both DeepDTI-WLLS and DoDTI outperformed MPPCA-WLLS, GLHOSVD-WLLS, and dtiRIM for FA and MD across all b values. Specifically, DeepDTI-WLLS slightly surpassed DoDTI for FA, while DoDTI performed marginally better for MD. Notably, even though trained on b = 1000 s/mm² volumes, both methods demonstrated good performance with b = 800 s/mm² and 1200 s/mm² datasets. Similar trends were observed for AD and RD metrics, as shown in Supplementary Fig. S1.

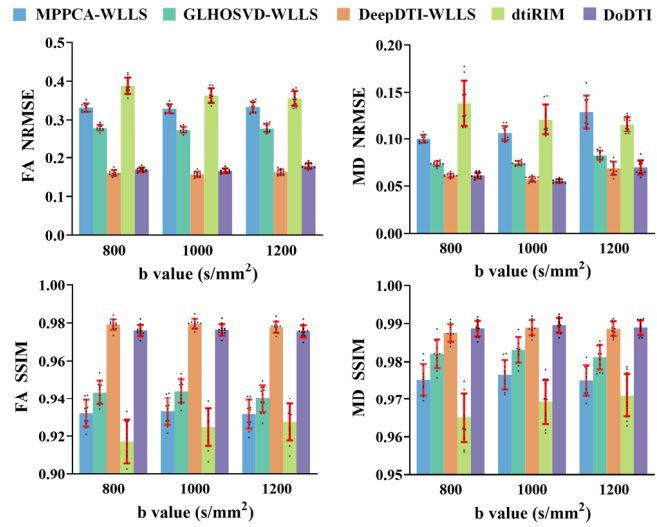

**Fig. 2.** Effect on different b values. Comparison of the NRMSE and SSIM of FA and MD estimated by different methods about a variety of b values.

#### 4.1.2 Effect on varying six directions

Fig. 3 illustrates the NRMSE and SSIM of FA and MD estimated by different deep learning-based methods, using datasets with six different diffusion directions. When the testing data aligned with the training data in diffusion directions (DSM scheme), DeepDTI-WLLS and DoDTI performed comparably, outperforming dtiRIM. However, as the discrepancy in diffusion directions increased between the training and testing datasets, DeepDTI's performance degraded, while DoDTI and dtiRIM maintained consistent results. Although dtiRIM exhibited robustness, its accuracy remained lower than that of DeepDTI and DoDTI. Similar trends were observed for AD and RD metrics, detailed in Supplementary Fig. S2.

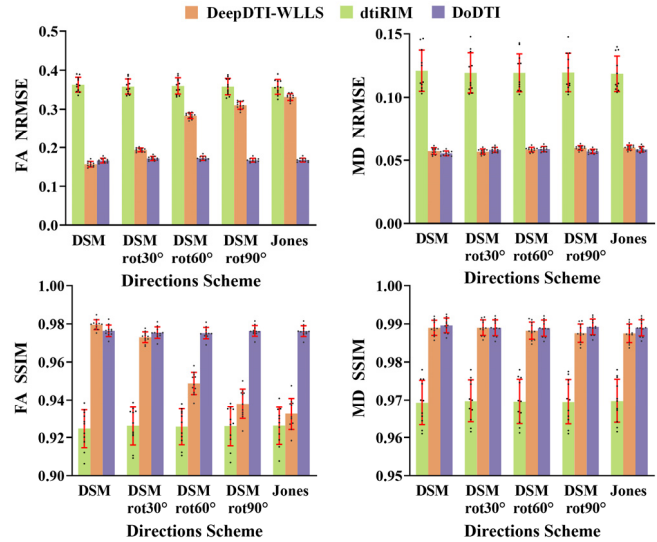

**Fig. 3.** Effect on varying six directions. Comparison of the NRMSE and SSIM of FA and MD estimated by deep learning-based methods about a variety of the six diffusion directions. DSM is the directions scheme of the training dataset. DSM rot30°, DSM rot60°, and DSM rot90° represent the DSM scheme rotated by 30, 60, and 90 degrees, respectively. Jones and DSM are an entirely distinct scheme with no rotational relationship.

### 4.1.3 Effect on different number of DW volumes

Fig. 4 presents the NRMSE and SSIM of FA and MD using datasets with varying numbers of DW volumes. DoDTI notably excelled MPPCA-WLLS, GLHOSVD-WLLS, and dtiRIM for both FA and MD across varying numbers of directions. Specifically, the results gradually declined as the number of directions decreased for the three methods, while the DoDTI produced a consistently accurate result even with seven DW volumes. Impressively, DoDTI achieved more accurate estimations with only seven DW volumes compared to other methods with 36 DW volumes. DeepDTI's limitation in processing only six DW volumes excluded it from this experiment. Similar trends were observed for AD and RD metrics, detailed in Supplementary Fig. S3.

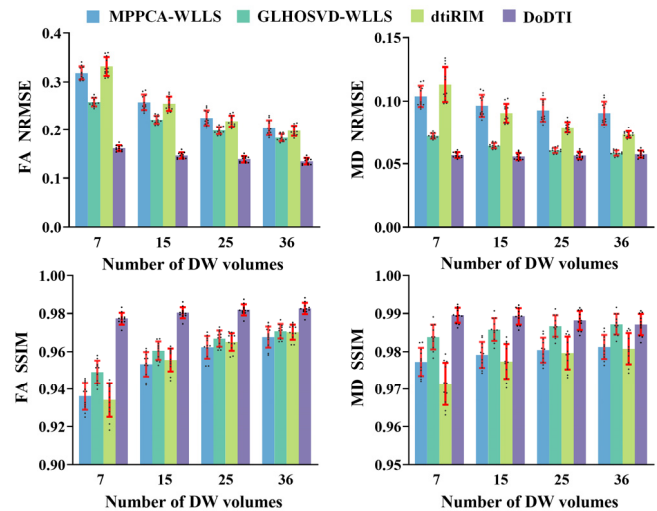

**Fig. 4.** Effect on different numbers of DW volumes. Comparison of the NRMSE and SSIM of FA and MD estimated by different methods about a variety of the numbers of DW volumes.

### 4.1.4 Effect on different noise levels

Fig. 5 exhibits the NRMSE and SSIM of FA and MD estimated by different methods, using the simulated datasets with stationary noise at different noise levels. Similar to the results of experiments on different b values, both DoDTI and DeepDTI-WLLS outperformed MPPCA-WLLS, GLHOSVD-WLLS, and dtiRIM across all noise levels, particularly at higher noise levels. Comparable results were noted for AD and RD metrics, outlined in Supplementary Fig. S4.

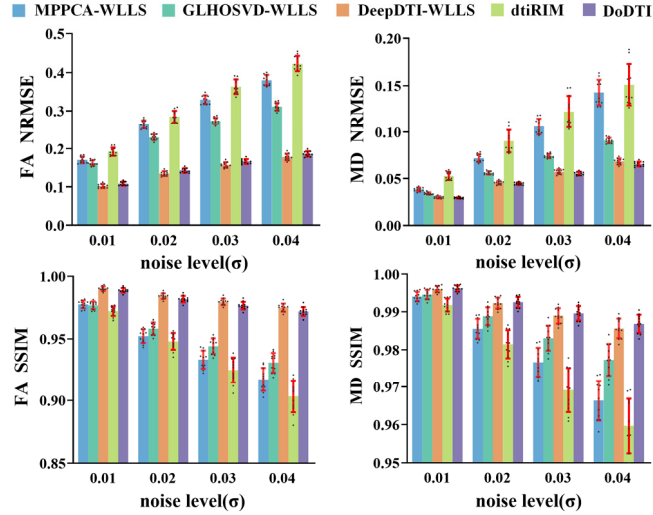

**Fig. 5.** Effect on different noise levels. Comparison of the NRMSE and SSIM of FA and MD maps estimated by different methods about a variety of noise levels.

### 4.1.5 Effect on spatially varying noise

Fig. 6 displays NRMSE and SSIM values for FA and MD estimated using various methods on the simulated dataset with spatially varying noise. Both DeepDTI-WLLS and DoDTI exhibited comparable performance across FA and MD, showcasing notably lower NRMSE and higher SSIM compared to MPPCA-WLLS, GLHOSVD-WLLS, and dtiRIM. The performance of MPPCA-WLLS, GLHOSVD-WLLS, and dtiRIM heavily relies on the accurate estimation of noise levels, proving challenging in the presence of spatially varying noise. Surprisingly, despite being trained under stationary noise conditions, DoDTI and DeepDTI-WLLS still demonstrated excellent performance in the presence of spatially varying noise. Similar findings were noted for AD and RD metrics, as outlined in Supplementary Fig. S5.

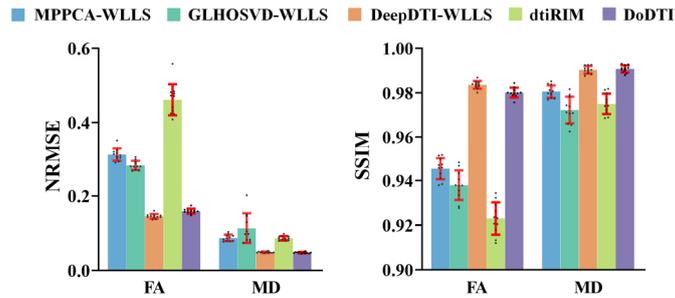

**Fig. 6.** Effect on spatially varying noise. Comparison of the NRMSE and SSIM of FA, MD, AD, and RD maps estimated by different methods about spatially varying Rician noise.

**4.1.6 Intermediate results of the iterative stages**

Fig. 7 illustrates intermediate results of DoDTI using simulated data with default settings. These results ($X^n$, $\mathcal{D}_\theta(Z^{n-1})$, and $Z^n$) showcased reduced noise and enhanced detailed information across stages, particularly in the initial four stages. The final estimated map ($X^8$, $N_s = 8$) was visually closed to the reference. Additionally, the CNN-based denoiser effectively eliminated noise at each stage. Quantitatively, RMSE/SSIM firstly rapidly decreased/increased and then gradually converged at a stable value with stages.

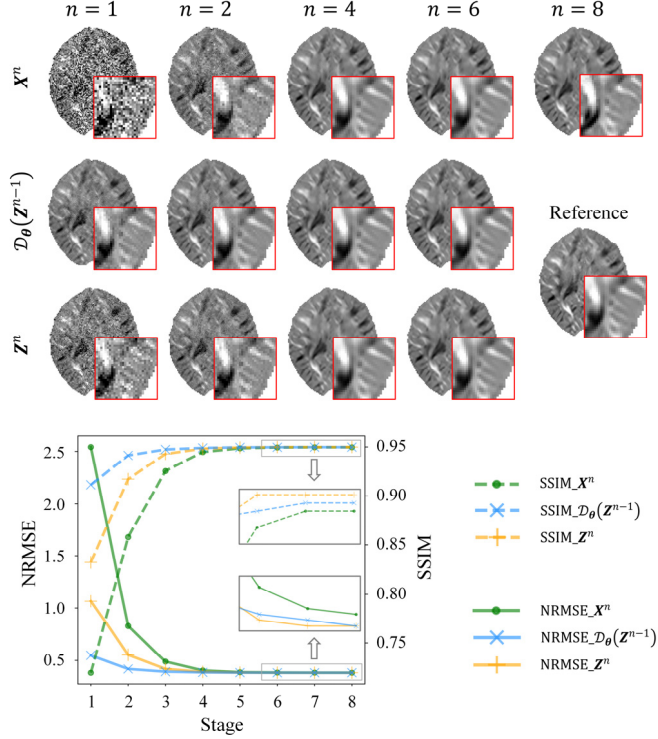

**Fig. 7.** Intermediate results in the deep unrolling network. At the *n*-th stage, $X^n$ is the output of the fitting block, $\mathcal{D}_\theta(Z^{n-1})$ is the output of the denoiser $\mathcal{D}_\theta(\cdot)$, $Z^n$ is the output of the auxiliary variable block. The representative maps ($D_{12}$) at stages 1, 2, 4, 6, and 8 are shown, in which $X^8$ is the final output of the network. The line graph shows the curves of NRMSE and SSIM versus stages for $X^n$, $\mathcal{D}_\theta(Z^{n-1})$, and $Z^n$.

**4.2 In-vivo Data Experiments**

Fig. 8 presents FA maps estimated by various methods using 6, 20, and 30 DW volumes from a healthy volunteer. DoDTI consistently produced less noisy FA maps with preserved detailed information, most closely resembling the reference across different DW volumes compared to other methods. Notably, WLLS performed poorly due to substantial noise impact, particularly with six DW volumes. MPPCA-WLLS and dtiRIM exhibited insufficient denoising, especially with fewer DW volumes, while GLHOSVD-WLLS yielded over-smoothed results. DeepDTI-WLLS produced a slightly blurred FA map due to differences between the real data's six sampling directions and the training data. Quantitatively, the DoDTI consistently derived the lowest NRMSEs and the highest SSIMs for FA maps across different numbers of DW volumes.

Moving beyond DTI parameter maps, we analyzed tractography results utilizing only six DW volumes. Fig. 9 illustrates a whole-brain white matter fiber group (comprising 10 major fiber tracts) as well as the cingulum cingulate fibers (CC) and inferior longitudinal fibers (ILF). As seen from the fiber group (row a), all methods successfully extracted most of the 10 fiber tracts using the limited six DW volumes, displaying visually similar reconstructed fibers without significant disparities. However, upon closer inspection of single fiber tracts (rows b and c), DoDTI notably identified more fibers of CC and ILF, closely resembling the reference. Specifically, the red arrows indicate increased cingulum fibers connecting the frontal, parietal, and temporal lobes in the DoDTI results.

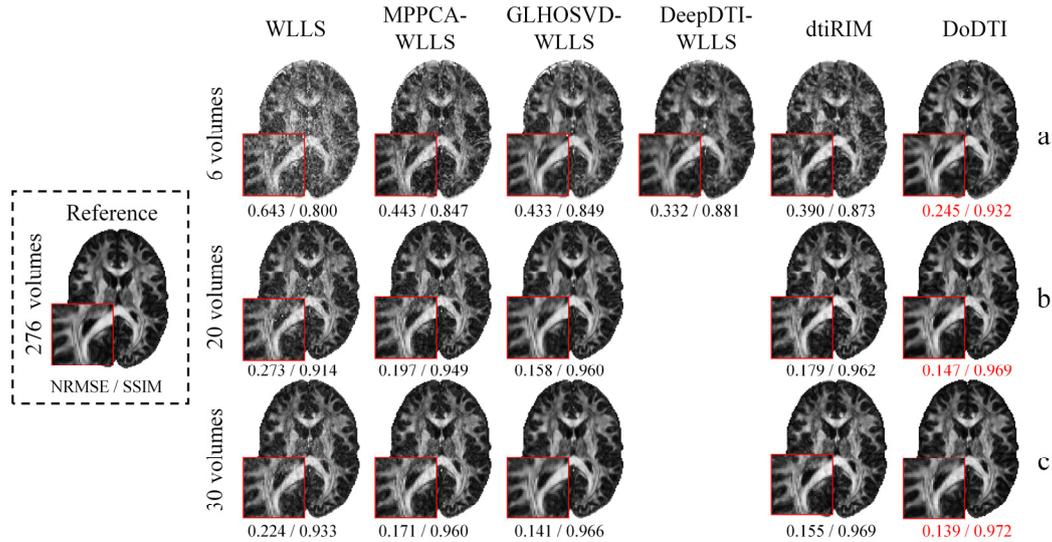

**Fig. 8.** FA maps estimated on different number of DW volumes using healthy volunteer data. The reference was estimated using WLLS from 276 DW volumes (46 DW volumes with 6 repeated scans). The FA maps were estimated using the 6, 20, and 30 DW volumes subsampled from the 46 DW volumes at the first scan. A magnified image is shown in the red box. The NRMSE and SSIM values are shown at the bottom of each magnified image. DeepDTI processes only six DW volumes.

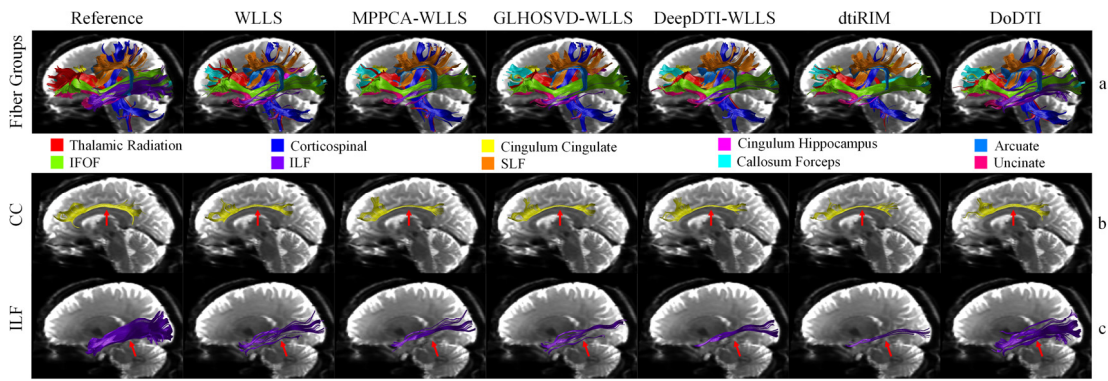

**Fig. 9.** Fiber tracking on six DW volumes using healthy volunteer data. The reference tracts were obtained from 276 DW volumes. The fiber group, as well as the cingulum cingulate fasciculus (CC) and inferior longitudinal fasciculus (ILF), are separately shown in rows a, b, and c.

Fig. 10 juxtaposes the FA and MD maps utilizing white matter hyperintensities data with 1.5 and 2 mm isotropic resolution, respectively. Across both resolutions, the FA and MD maps obtained from DoDTI exhibit markedly reduced visual noise compared to those derived from WLLS, MPPCA-WLLS, and dtiRIM. Additionally, they contain more nuanced structural details in contrast to the ones from DeepDTI-WLLS and GLHOSVD-WLLS. Notably, the FA and MD maps generated by DoDTI closely resemble the reference, showcasing the lowest NRMSE and highest SSIM. Remarkably, DoDTI achieves effective noise reduction without compromising the structure of small lesions—a crucial focus in clinical analysis, as shown in the enlarged views. The results from a patient post-glioma resection surgery with 1.5 and 2 mm isotropic resolution was shown in Supplementary Fig. S6. Similar to white matter hyperintensity data, DoDTI consistently outperformed other methods, suggesting its ability to generalize well in the absence of brain tissue structure.

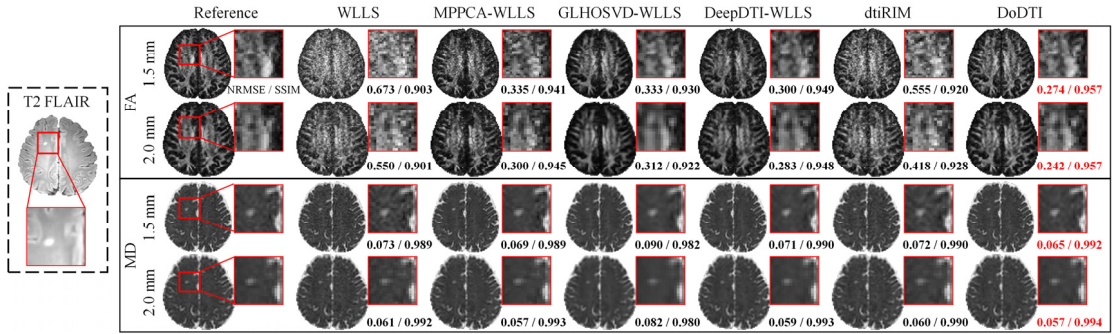

**Fig. 10.** FA and MD estimated using white matter hyperintensity data. The FA and MD were estimated using 6 DW volumes with 1.5 and 2 mm isotropic resolution, respectively. The reference was estimated using 64 DW volumes. A magnified view of the lesion is shown in the red box. The NRMSE and SSIM are shown at the bottom of each map.

Fig. 11 compares the FA and MD maps estimated by different methods using 6 DW volumes and 32 DW volumes from a glioma patient, respectively. Notably, the $R^2$ values of FA maps were consistently lower than those of MD maps for each method, suggesting that the FA map is more sensitive to noise and requires additional DW volumes to counteract noise effects. Among the methods, DoDTI produced the highest $R^2$ value for FA, while GLHOSVD-WLLS produced the highest $R^2$ value for MD. It's worth noting that the $R^2$ value of DoDTI (0.920) is only marginally lower than that of GLHOSVD-WLLS (0.992) for the MD map. Qualitatively, the FA maps estimated from 32 DW volumes displayed significantly reduced noise compared to those from six DW volumes for WLLS, MPPCA-WLLS, and dtiRIM. Conversely, FA maps were visually more consistent for DoDTI, particularly evident in the enlarged views. All methods produced relatively more consistent results for MD than for FA. Moreover, the glioma structure exhibited notably different characteristics from the training data, underscoring DoDTI's ability to generalize effectively to previously unseen anatomies. It is important to note that DeepDTI, limited to six DW volumes, was not part of this particular experiment.

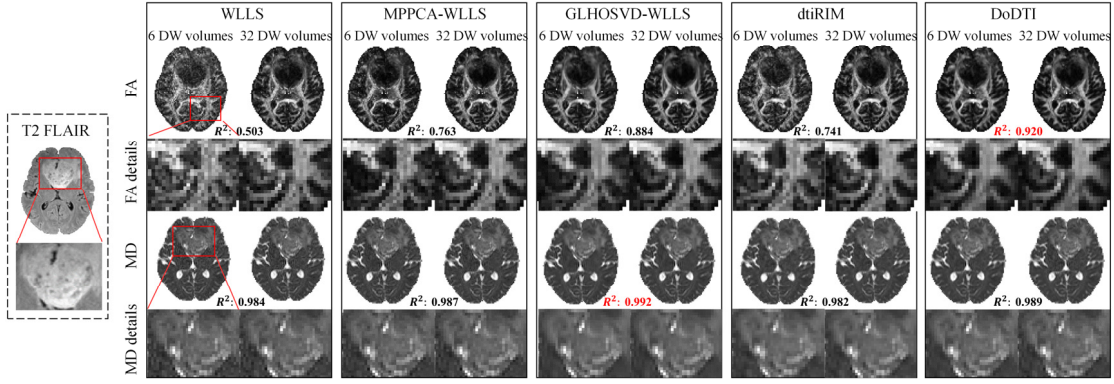

**Fig. 11.** FA and MD estimated using 6 and 32 DW volumes of glioma patient data. The magnified FA image highlights the detailed structure of the white matter, while the magnified MD image delineates the glioma region. The $R^2$ is shown at the top of each magnified map.

## 5. Discussion

Traditional model-based fitting methods in a voxel-wise manner are widely used in DTI parameter estimation due to their high robustness and ease of use. Nonetheless, these methods are highly sensitive to noise, resulting in poor accuracy, especially when the number of DW images is limited. Conversely, data-driven deep learning methods can achieve superior accuracy with fewer DW images, but they often lack generalization abilities (Monga et al., 2021), limiting their utility in neuroscience and clinical applications. To tackle this challenge, we rethought the power of the data-driven optimization routine and introduced a reliable deep learning-based method for DTI parameter estimation, effectively combining the advantages of conventional model-based fitting and deep learning methods. The results demonstrated that our method exhibited exceptional performance in generalization, accuracy, and efficiency.

In practical scanning scenarios, DW image acquisition protocols vary significantly among different centers, scanners, and studies. The purely data-driven deep learning methods such as DeepDTI (Tian et al., 2020) and SuperDTI (Li et al., 2021) are highly reliant on the number of input DW images and often necessitate retraining for datasets with different number of directions. Even when maintaining six DW images as input, altering gradient acquisition angles adversely affects their performance, such as DeepDTI (Fig. 2). This limitation is attributed to a "black-box" network structure that neglects diffusion gradient information, resulting in poor generalization and interpretability. By contrast, the strong generalization ability of the proposed DoDTI method is attributed to the following two aspects: (1) Incorporating the DTI model (including the strength and directions of diffusion gradients) in the data fidelity term, ensuring its generalization to the datasets with varying b values, different gradient directions, and different numbers of DW images; (2) By contrast, we use neural networks to directly regularize the parameter maps, which had a fixed number of network inputs, ensuring independence from DW image distribution and counts.

Apart from achieving generalization, researchers perpetually pursue highly accurate parameter imaging. While the dtiRIM, another data-driven model-based method, demonstrates good generalization ability, its competitive advantage in accuracy, particularly with fewer DW images, is lacking compared to other advanced methods (Fig. 4 and Fig. 8). Contrarily, in addition to its robust generalization, the proposed DoDTI method outperforms all compared methods in terms of accuracy.

NLS has been developed as an optimal estimator for the Gaussian noise model, assuming fitting error values ($\epsilon$) that allow a Gaussian distribution with zero mean and constant variance. However, DW images affected by noise follow a Rician distribution, causing NLS to theoretically bias tensor estimation. The WLLS corrects this bias of $\epsilon$ by using log-transformed signals and introduces a weighting factor to address the heterogeneity of variance of $\epsilon$. Consequently, it brings the mean of the error closer to zero and maintains a constant variance, approximating the fitting error values akin to a Gaussian noise model (Salvador et al., 2005; Veraart et al., 2013). Moreover, WLLS employs a linear formulation, sidestepping the complexity of nonlinear computations, and offers an analytical solution, circumventing the need for inner iterative solutions within the fitting block. Hence, in both theory and practice, we opted for WLLS over NLS as the data fidelity term in DoDTI. Notably, in clinical practice, the observed data may deviate from a strict Rician distribution due to factors such as partial Fourier sampling, GRAPPA reconstruction, coil correlations, and multi-band techniques. Nevertheless, DoDTI performed admirably with in-vivo data encompassing healthy and patient subjects across various scanners and centers.

In terms of computational efficiency, we compared the running speeds among different methods. To be fair, all methods were run in CPU mode. For dMRI data including one non-DW volume and six DW volumes (each volume sized 140 × 140 × 96), DoDTI and WLLS take approximately 1 minute, whereas DeepDTI-WLLS, MPPCA-WLLS, dtiRIM, and GLHOSVD-WLLS take approximately 1.5, 3, 6, and 50 minutes, respectively. In GPU mode, DoDTI only takes 16 s, highlighting its superior computational efficiency.

Distinguished from traditional optimization method, our approach introduces learnable parameters $\lambda$, $\rho$, and $\boldsymbol{\theta}$ as part of the optimization objective. These parameters are obtained via end-to-end training and are shared across all stages. Notably, the convergence may be enhanced if the network carries over layer-specific parameters ($\lambda$, $\rho$, and $\boldsymbol{\theta}$) (Chan et al., 2017), but the number of parameters will grow linearly with iterations. In addition, CNN weights $\boldsymbol{\theta}$ are customized to the forward model and learned through end-to-end training, offering faster convergence than traditional model-based methods (Aggarwal et al., 2019). Utilizing a pre-trained denoiser can further reduce training costs and memory requirements (Schlemper et al., 2018). Additionally, exploring more advanced network architectures of the denoiser might further enhance performance. Additionally, investigating the effect of the total number of stages ($N_s$) and the number of inner iterations in auxiliary variable blocks ($N_t$) on model performance is crucial. Currently, these hyperparameters are manually selected by experience, achieving a satisfactory convergence (Fig. 7). Automated hyperparameter tuning using Bayesian optimization or reinforcement learning could enhance model performance—an avenue worth pursuing.

While our evaluation of the DoDTI focused on DTI parameter estimation, its potential extends readily to other diffusion imaging techniques, such as DKI and NODDI, by modifying the physical model and the structure of denoiser.

## 6. Conclusion

This study addresses the limitations of current DTI parameter estimation by re-evaluating the potential of data-driven optimization strategies. We redefined the conventional optimization objective by introducing the traditional model-based fitting method and deep learning techniques. The combination of their respective advantages effectively overcomes the generalization weaknesses seen in purely data-driven neural networks. Moreover, it addresses the issues of slow arithmetic speed and noise sensitivity associated with model-based fitting methods. As a result, our deep DTI parameter estimation method exhibits substantial potential for broader applications in clinics and neuroscience, advancing the field of accelerated DTI imaging.

**CRediT authorship contribution statement**

**Jialong Li:** Conceptualization, Methodology, Validation, Visualization, Writing-original draft, Writing-review & editing. **Zhicheng Zhang:** Conceptualization, Methodology, Validation, Writing-original draft, Writing-review & editing. **Yunwei Chen:** Investigation, Data Curation, Validation. **Qiqi Lu:** Methodology, Investigation, Writing-review & editing. **Ye Wu:** Investigation, Resources, Formal analysis. **Xiaoming Liu:** Investigation, Resources, Formal analysis. **Qianjin Feng:** Supervision, Resources, Writing-review & editing. **Yanqiu Feng:** Conceptualization, Supervision, Writing-review & editing, Project administration, Funding acquisition. **Xinyuan Zhang:** Conceptualization, Supervision, Writing-original draft, Writing-review & editing, Project administration, Funding acquisition.

**Declaration of interest**

The authors declare no conflicts of interest.

**Acknowledgments**

This work was supported by the Natural Science Foundation of Guangdong Province (2023A1515012093), National Natural Science Foundation of China (U21A6005, 82372079, 61971214), and Guangdong-Hong Kong Joint Laboratory for Psychiatric Disorders (2023B1212120004). Data for simulation were provided by the Human Connectome Project, MGH-USC Consortium (Principal Investigators: Bruce R. Rosen, Arthur W. Toga and Van Wedeen; U01MH093765) funded by the NIH Blueprint Initiative for Neuroscience Research grant; the National Institutes of Health grant P41EB015896; and the Instrumentation Grants S10RR023043, 1S10RR023401, 1S10RR019307. The first in-vivo dataset was provided by Steen Moeller from the University of Minnesota. We acknowledge their contribution to our research.